\documentclass{article}
\usepackage{ijcai23}

\usepackage{times}
\usepackage{soul}
\usepackage{url}
\usepackage[hidelinks]{hyperref}
\usepackage[utf8]{inputenc}
\usepackage[small]{caption}
\usepackage{graphicx}
\usepackage{amsmath}
\usepackage{amsthm}
\usepackage{booktabs}
\usepackage{algorithm}
\usepackage{algorithmic}
\usepackage[switch]{lineno}

\usepackage{xcolor}
\usepackage{amsfonts}
\usepackage{array, comment}
\usepackage{mdframed}
\usepackage{lipsum}
\usepackage{upgreek}
\usepackage{makecell}

\title{Spectral 3D Computer Vision - A Review}

\author{
Yajie Sun$^1$
\and
Ali Zia$^{2,3}$\and
Vivien Rolland$^{2}$\and 
Charissa Yu$^{1}$\and
Jun Zhou$^1$
\affiliations
$^1$School of Information and Communication Technology, Griffith University, Australia\\
$^2$CSIRO Agriculture and Food, Australia\\
$^3$College of Science, The Australian National University, Australia
\emails
\{yanjie.sun, charissa.yu, jun.zhou\}@griffith.edu.au,
\{ali.zia, vivien.rolland\}@csiro.au
}

\begin{document}

\maketitle

\begin{abstract}
Spectral 3D computer vision examines both the geometric and spectral properties of objects. It provides a deeper understanding of an object's physical properties by providing information from narrow bands in various regions of the electromagnetic spectrum.
Mapping the spectral information onto the 3D model reveals changes in the spectra-structure space or enhances 3D representations with properties such as reflectance, chromatic aberration, and varying defocus blur. This emerging paradigm advances traditional computer vision and opens new avenues of research in 3D structure, depth estimation, motion analysis, and more. It has found applications in areas such as smart agriculture, environment monitoring, building inspection, geological exploration, and digital cultural heritage records. This survey offers a comprehensive overview of spectral 3D computer vision, including a unified taxonomy of methods, key application areas, and future challenges and prospects.
\end{abstract}

\section{Introduction}

3D computer vision is a challenging field of research that involves not only understanding the geometry and depth of a scene but also accurately representing its photometric characteristics. Historically, this information has been gathered and analyzed using grayscale or color images and traditional sensing technologies such as stereo vision, structured light, and LiDAR-based systems. However, these approaches often focus on information in the RGB domain and do not take advantage of the additional information that can be found in other wavelengths of the light spectrum like ultraviolet and infrared, or higher spectral resolution images like multispectral and hyperspectral.

The use of spectra of different wavelengths and fine spectral bands in 3D computer vision allows for a more comprehensive understanding of the surface of an object. Spectral information can accurately represent the inherent physical properties of an object, leading to a broader range of possibilities for obtaining high-quality 3D information. In recent years, there has been a growing interest in combining 3D computer vision with spectral analysis beyond the visible spectrum - which we refer to as {\bf Spectral 3D Computer Vision}. Many researchers have developed spectral 3D models to improve the accuracy and reliability of computer vision tasks in various applications such as environmental and plant modeling~\cite{liang20133d}, agriculture surveillance~\cite{padua2019vineyard}, digital cultural heritage records~\cite{chane2013registration}, and material classification~\cite{liang2014remote}.

To fully understand the current state of this area and identify future research directions, it is important to conduct a thorough review of the existing literature. However, most existing surveys have focused solely on conventional 3D computer vision with RGB data. Additionally, while some reviews have touched on relevant approaches involving spectra, they have been limited to specific topics such as agriculture remote sensing~\cite{jurado2022remote} and plant phenotyping~\cite{liu2020hyperspectral}. 

The focus of the review is on how 3D computer vision can be improved by utilizing spectral information. To the best of our knowledge, this is the first comprehensive survey on this topic. The aim of this survey is to help researchers understand the current state of the art, identify challenges, and provide guidance on future research.

The rest of this survey is organized as follows. In Section~\ref{sec:data}, we provide background information regarding 3D data, spectral data, and their acquisition. Section~\ref{sec:method} summarizes different spectral 3D computer vision methods, including multi-source data integration, structure from spectra, depth estimation, and pose and motion analysis. Section~\ref{sec:app} explores various applications, common tools, and libraries for spectral 3D computer vision. In Section~\ref{sec:discussion}, we discuss the limitations and opportunities of this field. Finally, we conclude the survey in Section~\ref{sec:conclusions}.

\section{Spectral 3D Data and Acquisition}\label{sec:data}

This section provides an overview of the fundamentals of 3D and spectral data, as well as their acquisition under controlled conditions and in natural environments. It also highlights some of the key challenges with respect to spectral 3D computer vision.

\begin{figure*}[ht]
\centerline{\includegraphics[width=0.96\textwidth]{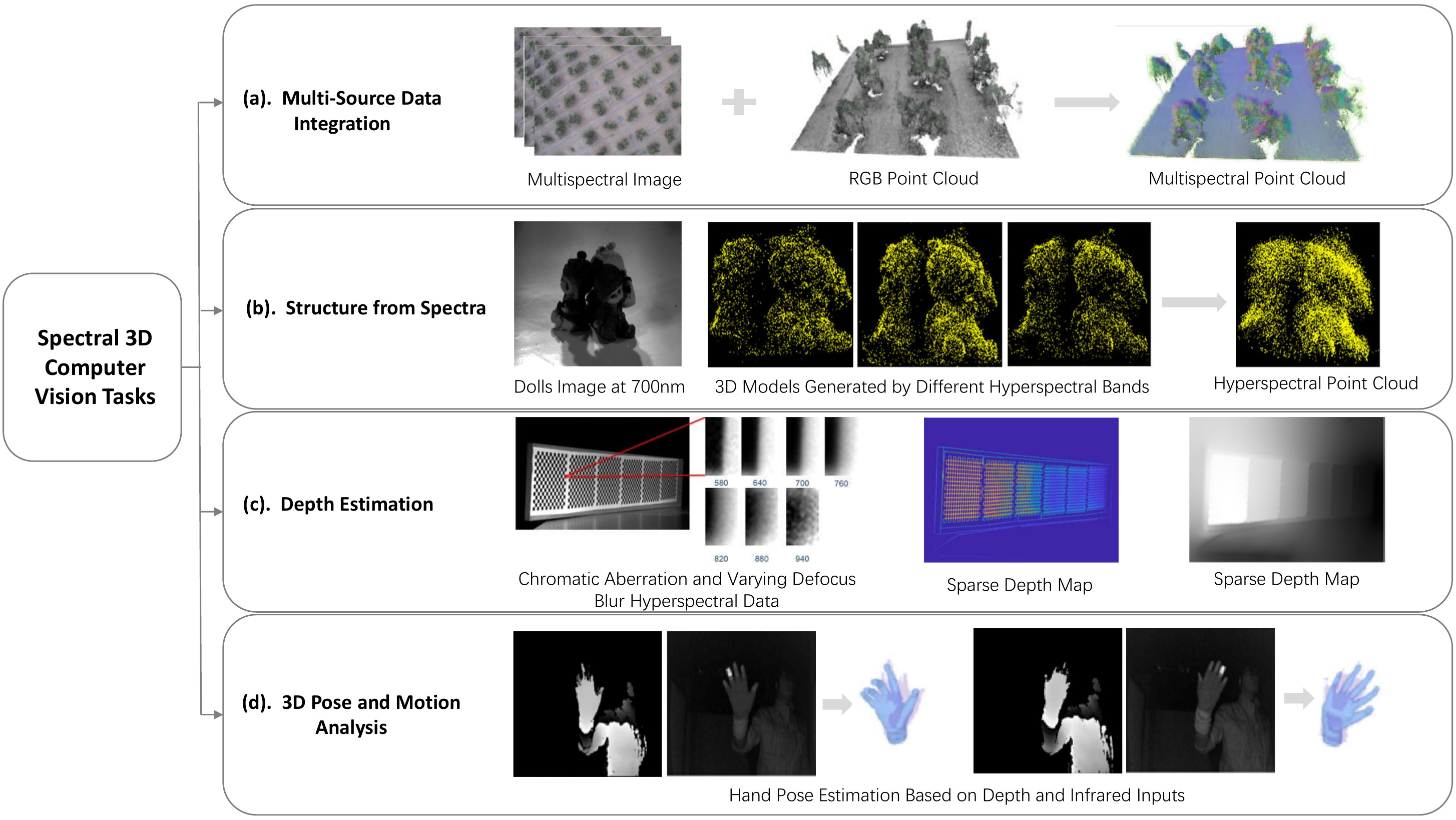}}
\caption{Spectral 3D Computer vision tasks. Process of (a) Multi-source data integration adapted from {\protect\cite{lopez2022generation}}; (b) Structure from spectra adapted from {\protect\cite{zia20153d}}; (c) Depth estimation adapted from {\protect\cite{zia2021exploring}}; and (d) 3D Pose and Motion Analysis adapted from {\protect\cite{park20203d}}.}
\label{fig:tasks}
\end{figure*}

\subsection{3D Data}

Obtaining 3D data can be complicated and may require measurement of camera positions or manual alignment of partial views of a scene. The approach and underlying task can greatly affect the complexity of this process. One way to acquire 3D data is by using 2D images captured by sensors and applying techniques such as structure from motion (SfM)~\cite{schonberger2016structure}, multi-view stereo (MVS)~\cite{li2021spectral}, and structured light~\cite{li2019pro}. Other methods include using LiDAR-based scanners~\cite{barton2021extending} or depth-sensing cameras~\cite{maier2012real}. 

When capturing 3D data it is important to consider the resolution of the sensors being used. Often, a LiDAR and an imaging sensor have different resolutions, which can create registration issues.

Because 3D data representation is not standardized, objects can be represented in the form of voxels, depth maps, point clouds, and meshes. This creates challenges when trying to incorporate additional information with 3D data.


\subsection{Spectral Data}

In computer vision, spectral data can be categorized by the range of its wavelengths and be defined as ultraviolet (10nm-400nm), visible (400nm-700nm), and infrared (700nm-1mm). The infrared spectrum can be further divided into near-infrared, short-wavelength infrared, medium-wavelength infrared, long-wavelength infrared, and far infrared. Long wavelength infrared is also called thermal infrared because it offers information on temperature and emission properties~\cite{avdelidis2003emissivity}. Depending on the spectral resolution, spectral data can be categorized as multispectral or hyperspectral. A multispectral image consists of discrete band images (usually at least 3) each covering a wide range of the spectrum. On the contrary, hyperspectral images (HSIs) normally contain tens or hundreds of consecutive narrow spectral band images, indexed by wavelengths~\cite{liang20133d}.

Multispectral and hyperspectral cameras require significantly more light (intensity) than conventional RGB cameras to generate images with a high signal-to-noise ratio. This is because the light received is divided into many bands; if the total light intensity is low, each band will receive a weak signal. This is an important consideration, especially for indoor setups. HSI obtained with insufficient illumination and intensity may be unable to embed accurate spectral properties in a scene, which in turn can affect the performance of spectral 3D computer vision tasks like SfM and depth estimation. Another important factor to keep in mind is the need for a relatively homogeneous light source, i.e. providing a fairly uniform emission over the wavelength range of interest. For example, sunlight is the ideal light source for the visible to the near-infrared range, but it is obviously not usable in indoor tasks. A black-body radiation source, such as a tungsten or halogen light source, has instead been widely employed as a substitute. However, it too has a relatively weak response in the blue region of its spectra.



\section{Spectral 3D Computer Vision Tasks}\label{sec:method}

The acquisition of 3D information is essential for many computer vision tasks, such as 3D structure reconstruction, 3D pose estimation, and 2.5D depth estimation. Besides geometry and color, spectral reflectance and temperature are important attributes to enrich 3D tasks~\cite{regaieg2021assessing}. Using data of different wavelengths, it is possible to represent target objects with comprehensive and complementary information beyond the capability of traditional 3D vision~\cite{hardeberg2001acquisition}. 
In this section, we provide a review of various spectral 3D methods for typical 3D tasks. Visual illustrations of some spectral 3D computer vision tasks are shown in Figure~\ref{fig:tasks}. 

\subsection{Multi-Source Data Integration}\label{sec:multisource}
Multi-source data fusion explores data with different properties to build high-quality 3D models. Using spectral information instead of RGB data allows for greater accuracy when handling objects with similar color and texture or with occlusions.

Two different types of integration methods are reviewed in this section: mapping and estimation. Mapping methods project pixels from a 2D spectral image to a 3D point cloud. Estimation methods utilize active illumination to capture spectral information and then estimate the spectral reflection of each point on the reconstructed 3D structure.

\subsubsection{Mapping Spectral Information to 3D Models}\label{sec:mapping}

As mentioned previously, spectral images and 3D data are often generated independently using different sensors. To this end, mapping is one of the key methods for multi-source 3D data fusion. In general, the idea is to project pixels from 2D spectral images to points in a 3D point cloud. 

Let $\mathbf{H} \in \mathbb{R}^{N \times B}$ be a spectral image, where $N$ is the number of data samples and $B$ is the number of bands. Note that for single-band images like ultraviolet or infrared, $B=1$. Let $\mathbf{H}_g$ be a group of spectral images where $g$ is the number of images in the group. $\mathbf{P} \in \mathbb{R}^{M \times 3}$ denotes a point cloud, where $M$ is the number of points. The complete spectral 3D model $\mathbf{D}_{spec}$ can be expressed as:
\begin{equation}
\mathbf{D}_{spec}= \mathbf{H}_g \Pi \mathbf{P}
\end{equation}
where $\Pi$ is a projection operator determined by different mechanisms. 

The simplest method to map spectral data with a point cloud is to use a time-synchronized system~\cite{Elbahnasawy2018}. $\Pi$, in this case, is a function of time, where projection from the imaging sensor is time synchronized with the point cloud being generated. The camera needs to be calibrated and synchronized with the LiDAR unit to give accurate point-wise mapping. Furthermore, in the case of remote sensing additional georeferencing like Global Navigation Satellite Systems (GNSS) and Inertial Navigation Systems (INS) shall be used to provide accurate location information for better registration.

Time-synchronized calibrated systems are generally expensive and prone to errors derived from wrong calibration and low LiDAR resolution. This issue can be resolved by adopting the classic SfM approach~\cite{lopez2022generation}. SfM is a photogrammetric range imaging method for estimating three-dimensional structures from two-dimensional image sequences that are coupled with the local movement of the camera. To this end, hyperspectral images shall be first rectified and aligned with RGB images. The RGB point cloud is then generated from SfM, so that hyperspectral images can be projected onto the RGB point cloud. During this task, different optimization techniques~\cite{Graciano2021}
can be adapted to address compression and color aggregation issues. ~\cite{lopez2021optimized} followed a similar procedure but used an Enhanced Correlation Coefficient (ECC) algorithm to register the RGB and thermal-infrared multispectral images. Here $\pi$ is an aggregating operator encapsulating certification, alignment, and projection operations.

A common challenge faced in the mapping methods is the high geometric deformation of the spectral images. ~\cite{jurado2020multispectral} proposed an approach to incorporate multispectral images and high-resolution point clouds for plant monitoring. Since multispectral images are highly deformable, the inverse 3D projection was developed to assign each 3D point of a point cloud to a corresponding pixel in a multispectral image.

\subsubsection{Spectral 3D Model Estimation}\label{sec:modelEstimation}
In the absence of professional spectral cameras, photometric representation and 3D reconstruction from multiview spectral images require light sources to obtain spectral and geometric information. In a system for acquiring multiband and multiview images, let $L_s$ be a light source, $Cam$ be the camera and $\Psi$ be the mechanism obtaining 3D mapped data from the system, a spectral 3D model can be generated as:
\begin{equation}
\mathbf{D}_{spec} = \Psi(L_s,Cam).
\end{equation}
In such systems, wide-spectrum grayscale cameras that are sensitive to all interested wavelength ranges are usually the preferred option. Therefore, the key to spectral 3D modeling is the choice of light source and mapping mechanism.

A light source that can project varying spectra is often used in multispectral data acquisition whereas multiview images are used to capture the shape of the object.
~\cite{ito20173d} used estimated spectral reflectance from signals captured by the imaging sensor using the Wiener estimation method - a filtering-based technique~\cite{tsuchida2013stereo}. Colors could be reproduced with the estimated spectral reflectance and used to further estimate the image texture of objects. This process also allowed true colors that reflect spectral information to be combined with 3D models by matching pixels in 2D images with points in 3D. 

However, a multispectral light source can be expensive and impractical for some applications. A viable option is to use an off-the-shelf projector to simulate a multispectral light source.
~\cite{li2019pro} proposed a method to estimate the spectral reflectance of each 3D point using a projector. This produced a 3D model of higher quality than one obtained from images captured by a monochrome or RGB camera. 
The parameters of this estimation model included the geometric relationship between the 3D point and the estimated projector position. 
Furthermore, both photometric and geometric observations can be obtained through a sequence of uniform colors and pattern projections, where uniform color illuminations can simulate multiple spectra, and  
structured patterns, like binary gray codes~\cite{geng2011structured} are used to get structural information. 

Inspired by the use of projectors as light sources, ~\cite{li2021spectral} utilized LED bulbs with different spectral power distributions to get multispectral data.
The geometric and photometric principles in image formation were considered in terms of camera spectral sensitivity, spectral power distribution of light, and light source location. Based on this, a multi-view inverse rendering framework was constructed to jointly reconstruct 3D shapes and vertex spectral reflectance while estimating light source position and shadow. Accurately estimating 3D points and spectral reflectance allows for spectral 3D relighting of objects to synthesize the appearance of illumination under an arbitrary light orientation and spectral distribution. As a result, high-quality 3D models with spectral information can be generated.

\subsection{Structure from Spectra}\label{sec:structure}
Although many studies have incorporated spectral data into 3D modeling, none of these methods directly reconstruct a 3D model from the spectral data. Given that the surface of the object may have different reflection characteristics due to complex material composition and that the change of wavelength causes a change of focus~\cite{garcia2000chromatic}, different spectral bands used to reconstruct 3D models show distinctive structural traits. If a single band or greyscale image derived from HSI is used for reconstruction, fine structural information might be lost. Therefore, following the standard SfM technique~\cite{schonberger2016structure}, 3D models can be generated band-by-band to obtain the 3D structure from multiview images of the same wavelength and then fused to form multispectral 3D models. 
 
~\cite{liang20133d} built a 3D model from different wavelengths of a hyperspectral image for the study of plant phenomics. In a controlled laboratory environment, multiview hyperspectral images were captured and their background was removed by SVM and K-means methods. 3D models from different wavelengths were then generated using the SfM approach. 
This work showed that each band contributes to different aspects of an object’s 3D structure. Obtaining a complete 3D reconstruction from this data would require registration of all band-wise 3D models. This is a difficult task because each 3D model has inconsistent scales and point sets. 

In order to maintain fine structure and spectral information embedded in the band image, various scale estimation methods have been proposed~\cite{ma2013efficient}. 
For example, \cite{zia20153d} reconstructed a complete hyperspectral 3D model by improving feature matching and point cloud registration. Unlike other methods based on surface norm~\cite{rusu2009fast}
, in \cite{zia20153d} Euclidean distances between points were calculated to generate a point-wise descriptor to characterize the statistical relationship of every single point to all other points in the 3D model. A novel registration method then integrated band-wise 3D models into a complete hyperspectral 3D model. This registered 3D point cloud was denser and more complete (i.e. missing fewer structural details) than the point cloud of an individual wavelength. This work was extended~\cite{Zia2018} to register low-spatial resolution hyperspectral plant models and further analyze structural information. A pan sharpening~\cite{Loncan2015PanSharpning} technique was used to integrate low-resolution spectral images with a high-resolution color or grayscale image to create a high-resolution hyperspectral image before 3D reconstruction. 
The higher spatial resolution image further enhanced the constructed 3D model. This work showed that a more accurate, complete, and denser 3D model can be obtained as the hyperspectral camera’s spatial and spectral resolution gets better. 

\subsection{Depth Estimation}\label{sec:depth}

Depth estimation can be utilized to generate a 3D model of a scene for applications like augmented reality, robotics, and autonomous vehicles. Spectral information such as reflectance, chromatic aberration, and varying defocus blur present in multispectral and hyperspectral images have the potential to give more cues about depth than conventional RGB images.

In dual camera systems, pixels from multiple bands (commonly in the form of reflectance or irradiance) help to get a better correlation for the stereo disparity. ~\cite{heide2018real} uses the spectral information in a 16-band hyperspectral image to calculate correlation-based similarity measures among stereo pairs in the matching process. This information is further used to perform a consistency check for disparity selection. In ~\cite{luo2017augmenting}, a CNN-based depth model trained on synthetic data is used in conjunction with a multispectral photometric stereo setup is used to estimate depth information. The setup consists of an orthographic camera and multispectral (three colored) light sources projected on objects from different angles.
In this method, a CNN was first used to estimate a depth approximation, and multi-spectral photometric stereo normals were then used to progressively refine the depth information.

Hyperspectral images can further contribute to depth estimation in the form of inherent optical phenomena such as varying defocus blur and chromatic aberration. These optical effects have been actively studied in monocular depth estimation. Using the chromatic aberration,~\cite{kumar2015defocus} and~\cite{ishihara2019depth} captured defocus blur at various wavelengths. Based on the estimation of defocus parameters, ~\cite{kumar2015defocus} demonstrated that depth maps obtained from multispectral component representation were smoother and more accurate than those obtained from grayscale representation. In~\cite{ishihara2019depth}, multiple spectral components were used as multiple focal lengths to estimate depth.

Furthermore, Zia et al. analyzed spectral chromatic and spatial defocus aberration in a monocular hyperspectral image to get a more accurate depth map than RGB-only approaches~\cite{zia2015relative,zia2021exploring}. They presented various methods such as manifold learning or customized optimization strategy to calculate and combine these properties, which generated optimal sparse depth maps. Then the dense depth map could be generated using different graph Laplacian approaches.
 
In summary, depth estimation using extra information in multispectral and hyperspectral imaging is still an emerging area. New optimized algorithms and hardware are likely to further refine depth estimation and boost spectral 3D vision.

\subsection{3D Pose and Motion Analysis}\label{sec:pose}

3D pose estimation can transform objects present in 2D images into 3D objects and has been widely adopted to predict and track the actual spatial orientation of objects. It enables higher performance and more effective representation of image features. However, RGB images are sensitive to environmental factors such as shadows and changes in lighting conditions, causing problems in pose estimation~\cite{schonauer20133d}.

Compared with RGB images, infrared (incl. thermal infrared) images are less susceptible to environmental factors such as changes in illumination. \cite{nishi2017use}, for example, used 3D and thermal information to identify humans based on their shape and to detect their location. Employing a thermal point cloud, which was generated by combining data from a calibrated pair of thermal and depth cameras, could effectively extract regions containing humans. In addition, the combination of infrared images and point clouds obtained from time-of-flight cameras have been used for 3D driver pose estimation~\cite{yao20203d} and could enhance the performance and speed of 3D human pose estimation.

Spectral 3D computer vision can be adapted to capture finer structures in pose analysis. \cite{park20203d} proposed the first solution for 3D hand bone recognition based on infrared data. This method used self-supervised domain transfer learning from a depth image to an infrared image, to accurately estimate 3D hand poses under motion blur caused by fast hand motion. In another example, point clouds generated by infrared images were used for pose estimation with Augmented Reality (AR) glasses ~\cite{firintepe2021ir}. Using 3D point clouds as an intermediate representation, two pose estimation algorithms for AR glasses based on single infrared images were proposed. The experiment proved that low-resolution point clouds generated from low-cost infrared hardware resulted in high pose estimation accuracy with excellent scalability. 

Further, pose estimation with 3D data and spectral information can also be applied in~\cite{firintepe2021ir}:
\begin{itemize}
    \item {\em Robotics:} Robots can assess their poses in the surrounding environment more precisely with 3D spectral data.
    \item {\em Motion Capture:} Motion capturing systems used in animation and film production can use 3D spectral data to better estimate pose and provide a more immersive experience for spectators. 
    \item {\em Medical surgical guidance:} More accurate pose estimation using spectral 3D data can be used to track the location and orientation of medical devices more precisely during surgical operations, thus lowering the risk of complications.
\end{itemize}

\section{Applications and Tools}\label{sec:app}
This section describes several application areas where spectral 3D computer vision has already contributed significantly. In addition, it introduces some tools and libraries that can help new users in this area.

\subsection{Applications}
Spectral 3D computer vision has been applied to a range of domains when traditional RGB-based 3D modeling could not provide sufficient information to accomplish certain tasks. Here we summarize relevant work performed in four representative areas.

\subsubsection{Smart Agriculture and Environmental Monitoring}
To monitor growth in vineyards, ~\cite{padua2019vineyard} used an unmanned aerial vehicle (UAV) to acquire RGB, multispectral, and thermal aerial images. The 3D point cloud with spectral features facilitated measuring soil moisture levels and detecting pests, diseases, and possible problems with irrigation equipment. 

In environmental monitoring, the fusion of hyperspectral and LiDAR helped to monitor forest diversity and structure~\cite{de2021monitoring,mayra2021tree}. LiDAR data described the height and size of trees, and the variability of spectral responses was related to species richness. Utilizing deep learning methods to extract spatial and spectral features from the data allowed high-precision species identification. This combination of information sources helped to understand changes in ecosystem composition, structure, and function of forests.



\subsubsection{Structural Inspection of Buildings}

Due to the diversity and complexity of building structures and materials, UAVs mounted with multiple sensors, collect visible and thermal-infrared data that is often used for building inspections ~\cite{carrio2016ubristes}. A thermal-infrared sensor can record and track the temperature, an object radiates to build global homogeneous models. 
This contributes to identifying architectural anomalies and defects. Typical applications include detecting earthquake-damaged buildings ~\cite{zhang2020automatic} or subsurface defects that cause heat loss ~\cite{puliti2021automated}. The combination of infrared information and 3D structures facilitates the segmentation of damage contours, remote inspection, and damage detection in large-scale systems.


\subsubsection{Natural Environmental Exploration}

Hyperspectral and LiDAR data have been used for 3D geological modeling. Reflectance and emittance spectra of natural surfaces measured by the hyperspectral sensor are sensitive to specific chemical bonds in materials, while LiDAR data conveys information on terrain and vegetation. \cite{nieto20103d} introduced a remote sensing system for mapping the geology and geometry of the environment. By combining hyperspectral information with 3D LiDAR data, a 3D map of the geological environment was created. The 3D map helped to estimate the grade and volume of ore being mined. Similarly, ~\cite{barton2021extending} integrated hyperspectral imaging techniques with LiDAR data to map the distribution of large-scale minerals. By analyzing hyperspectral images and 3D spatial georeferencing, vector data was generated to show mineral zonation and lithologic boundaries. In addition to geological exploration, spectral 3D information was also useful for mapping underwater environments ~\cite{ferrera2021hyperspectral} and contributed to a better understanding of the marine ecosystem.


\subsubsection{Digital Cultural Heritage Records}

Spectral and 3D imaging technologies have been used in museum imaging and cultural heritage records to provide supplementary information which aids in documenting artifact conditions, guiding care strategies, and increasing understanding of targets. ~\cite{chane2013registration} proposed a multi-sensor registration technology for 3D and multiple spectral data of cultural heritage value. ~\cite{grifoni2018construction} proposed a procedure for constructing and comparing 3D multi-band models obtained from a variety of input data and/or acquired with different instruments. It revealed sub-millimeter differences between 3D models generated by different spectral data sources and enabled the rapid delivery of digital products characterized by high levels of spectral and morphological details. In ~\cite{adamopoulos2020image}, image-based modeling with near-infrared images of heritage artifacts was evaluated. The results of this study were used to support conservation interventions.


\subsection{Tools and Libraries}

Over the past years, several tools have been developed for spectral 3D computer vision. Here, we summarize them to help interested researchers investigate this area.

\subsubsection{Spectral 3D Reconstruction}
Several publicly available tools can directly generate 3D models from a sequence of images. For example, Visual SFM\footnote{\url{http://ccwu.me/vsfm/index.html}} and COLMAP\footnote{\url{https://colmap.github.io/}} can automatically utilize SfM and MVS for 3D reconstruction without camera calibration. Additionally, open-source libraries such as Open3D\footnote{\url{http://www.open3d.org/}} and Spectral Python\footnote{\url{https://www.spectralpython.net/}} (SPy) provide support for processing 3D data and spectral images respectively. When generating 3D models from each wavelength, SPy can be used to load multiview hyperspectral images and perform prepossessing where needed. COLMAP can then be used to generate point clouds for each wavelength. Additional processing on the point cloud can be done using the Open3D library. Furthermore, libraries like PyTorch and TensorFlow can be used to learn a variety of tasks such as mapping, alignment, merging point clouds, etc at different stages of the pipeline.

\subsubsection{3D Model Visualization}
Open3D can be used to display 3D models. When further analyses or post-processing are required, applications like MeshLab\footnote{\url{https://www.meshlab.net/}}, CloudCompare\footnote{\url{https://www.danielgm.net/cc/}} and Blender\footnote{\url{https://www.blender.org/}} can be useful. They provide a graphical interface to display 3D point clouds and perform functionalities such as registration, alignment, filtering, smoothing, noise removal, texture mapping, etc.

\begin{table*}
	\centering
	\resizebox{1.0\textwidth}{!}{
    \begin{tabular}{|m{4cm}<{\centering}|m{6cm}|m{3cm}|m{4cm}|m{9cm}|}
		\hline
        \textbf{Task} & \centering{\textbf{Method}}   & \centering{\textbf{Platform}} & \centering{\textbf{Spectral Range}} & \centering\arraybackslash{\textbf{Evaluation Metric}}
        \\ \hline
		Mapping Spectral Information to 3D Models~\cite{jurado2020multispectral}  &  Iterative-closest-point (ICP) algorithm is used to map low-resolution spectral information to high-resolution RGB point cloud  & \makecell[l]{Multispectral Sensor, \\ RGB Camera, UAV}   & \makecell[l]{770-810nm, 730-740nm, \\ 640-680nm, 530-570nm} &  Root Mean Square Error(RMSE) for verifying geometric precision between the coordinates of reconstructed 3D points and their actual location.
        \\ \hline
		Mapping Spectral Information to 3D Models~\cite{lopez2021optimized}  &  Enhanced Correlation Coefficient (ECC) algorithm is used to project and fuse thermal infrared images on the 3D point cloud & \makecell[l]{Thermal Sensor, \\ RGB Camera, UAV}  & \makecell[l]{0.9$\upmu$m–1.7$\upmu$m, 3$\upmu$m-5$\upmu$m, \\ 8$\upmu$m–14$\upmu$m}  & RMSE, Mean Absolute Error(MAE), and Standard Deviation to evaluate the accuracy of aggregating spectral data projected on the RGB point cloud. 
        \\ \hline
        Spectral 3D Model Estimation~\cite{li2021spectral} & A inverse rendering approach is used to reconstruct the 3D shape and spectral reflectance jointly using multiview and multispectral images & \makecell[l]{RGB Camera, \\  LED Bulbs} & 400nm-700nm &  RMSE and metrics described in~\cite{Ley2016} to do a qualitative and quantitative evaluation for completeness and accuracy of the 3D structure and estimated reflectance.    
        \\ \hline
        Structure from Spectra~\cite{zia20153d} &  SfM to generate models and distance-based approach to register models & \makecell[l]{Hyperspectral Camera, \\ Halogen Lights}           & 400nm-700nm          & Mean Distance Error (MDE) to evaluate accuracy between proposed and other registration strategies on 3D models generated by spectral and RGB images.
        \\ \hline
        Depth Estimation~\cite{ishihara2019depth} & Derivative-based approach to exploit chromatic aberration present in a single multispectral image & Multispectral camera  & 450nm-700nm          & Mean Square Error(MSE) and Standard Deviation(SD) to evaluate the accuracy of the estimated depth with ground-truth data. 
        \\ \hline
        Depth Estimation~\cite{zia2021exploring} &  Manifold learning and graph Laplacian approaches to exploit chromatic aberration and defocus blur present in monocular hyperspectral images  & \makecell[l]{Hyperspectral Camera, \\ Halogen Lights}       & \makecell[l]{470nm-900nm (outdoor), \\ 590nm-920nm (indoor)}         & Visual and quantitative comparison of depth values in a depth map with those obtained by other methods. 
        \\ \hline 
        3D Pose and Motion Analysis~\cite{yao20203d} & Joint deep neural network architecture for registered 2D infrared images and 3D point clouds     & Infrared \& Point Cloud from Time-of-Flight Cameras           & 700nm-1mm         &  Average Error(AE) and Mean Average Precision(MAE) to evaluate the accuracy of prediction of pose estimation compared with ground-truth data and other methods.  
        \\ \hline
	\end{tabular}}
\caption{Comparison of selected spectral 3D computer vision methods for different tasks.}  
\label{tbl:comparission}
\end{table*}


\section{Discussion}\label{sec:discussion}
Spectral 3D computer vision is an active and promising area due to its flexibility in jointly handling spectral and structural information. Compared with most traditional computer vision data, the combined hyperspectral/multispectral and 3D data have higher dimensions. This leads to challenges in representing complex data and generalizing these representations for learning tasks. In addition, as an emerging area, spectral 3D computer vision does not have standardized datasets and evaluation matrices. In this section, we compare some methods from spectral 3D vision tasks and then discuss challenges and future directions. 
\subsection{Comparisons between spectral 3D vision tasks}
Table~\ref{tbl:comparission} compares sensors, spectral ranges, and evaluation metrics used in selected methods for spectral 3D computer vision. This table also shows that there lacks standardized evaluation criteria in this field, which makes it challenging to compare the performance of different methods. One reason for this issue is due to the fact that different hardware setups, wavelength ranges, and software pipelines have been used. Therefore, a possible effort is to identify the most successful and readily available setups and use them to collect data that could be made publicly available.

Evaluation of spectral 3D computer vision techniques can be done in a number of ways. One approach is to observe changes between specific wavelength bands. Another approach is to compare models generated with hyperspectral data with those obtained from high spatial resolution RGB data. Finally, as there are relatively few 3D spectral vision approaches that use a standardized setup, it is also possible to compare the results of hyperspectral methods with those obtained from their closest RGB counterpart. From the table and previous studies, the Root Mean Square Error (RMSE) is the most common metric because of its simplicity and flexibility, making it amenable to a variety of scenarios such as comparing individual 3D points or higher dimensional constructs.

\subsection{Challenges and Future Prospects}
Below, we highlight some gaps and promising future research directions which should motivate researchers to continue to explore new data, theories, and applications in spectral 3D computer vision. 

\textbf{Rise of computing-friendly devices:} With the mass of high-resolution spectral images and the huge amount of 3D points in a point cloud (often several hundred million points), the development of computational and storage capabilities, databases, and true collaborative environments is needed. Accordingly, effective methods must be developed to leverage hardware capabilities and ensure high computing performance. To deal with the challenge of processing huge amounts of input data from different sources, specific hardware, and algorithms are needed 
to take advantage of parallel and distributed computing strategies~\cite{jurado2022remote}. 

\textbf{Curse of dimensionality:} Compared to 2D plane data, 3D data has more spatial dimension and more complex structures. In addition, rich spectral information extends its characteristic dimensions. Although these properties are useful to get more accurate modeling outcomes, the curse of dimensionality is a critical problem with the increased dimensions. To minimize the impact of this curse, dimensionality reduction methods such as singular value decomposition can be used to simplify the data. 

\textbf{Insufficient public spectral 3D dataset:} Compared with common point cloud datasets, the spectral 3D data is insufficient. Even though some remote sensing datasets are publicly available, they are focused on specific tasks such as land cover classification or hyperspectral unmixing and are not always suitable for general computer vision tasks. This is mainly because spectral 3D vision is a relatively new area, and variability in data acquisition methods limits the availability of standardized benchmark datasets.
This lack of relevant public datasets has hindered the replication of algorithms and their performance evaluation. Therefore, collecting and publishing open spectral 3D datasets is an urgent task.


\textbf{Lack of spectral information:} In the process of 3D reconstruction, some images cannot generate a complete 3D model because the number of key points extracted from certain wavelengths or views is insufficient. Therefore, in the reconstruction experiment, some bands and views have to be removed. As a result, the final 3D model does not contain all the photometric information of objects. In future work, methods should be developed to reconstruct 3D models directly from the whole multispectral image rather than from multiple 3D band-wise models.

\textbf{Reflectance estimation with mutual reflection:} Current methods to estimate spectral reflectance consider the effects of cloud coverings and shadows but ignore the effects of mutual reflection. It is possible to resolve the limitations by estimating spectral reflectance and illumination spectra simultaneously or by separating the direct and global components using projected high-frequency illumination.

\textbf{Need for an underlining theoretical framework:} The spectral and structural properties of an object have different characteristics. In spectral 3D computer vision, the use of spectral-structural information is usually problem specific. Therefore 
a more general theoretical framework needs to be explored that addresses the underlining relationship between spectra and the structure of an object.

\textbf{More robust imaging setup:} Light sources are essential in hyperspectral imaging. Their settings determine the reflection path of the light on the surface of an object. Although hyperspectral imaging has high capability in distinguishing materials, a stable light source and reasonable illumination layout are key prerequisites for obtaining high-quality and high-stability images. An optimized diffuse light source system with intelligently-adjusted light intensity for hyperspectral imaging should be helpful to obtain hyperspectral images with good stability.


\section{Conclusion}\label{sec:conclusions}

Spectral information contributes to 3D computer vision in a variety of ways, e.g. estimating the structure of the object from spectra or getting more depth estimation cues in the underlining spectra. This review serves as a starting point to understand the definition of spectral 3D computer vision taxonomy, fundamental theories and methods to joint spectral-structural modeling, the usefulness of the technologies in various applications, and its associated challenges and future work.

\bibliographystyle{named}
\bibliography{ijcai}

\end{document}